\documentclass[runningheads]{llncs}
\usepackage[T1]{fontenc}
\usepackage{amsmath}
\usepackage{amssymb}
\usepackage{graphicx}
\usepackage{multirow}
\usepackage{caption}
\newcommand{\tabincell}[2]{\begin{tabular}{@{}#1@{}}#2\end{tabular}}
\begin{document}
\title{SAM++: Enhancing Anatomic Matching using Semantic Information and Structural Inference}
%
%
\author{Xiaoyu Bai; YongXia}
%
%
\institute{School of Computer Science and Engineering, Northwestern Polytecnical University
\email{bai.aa1234241@mail.nwpu.edu.cn}}
\maketitle              
\begin{abstract}
Medical images like CT and MRI provide detailed information about the internal structure of the body, and identifying key anatomical structures from these images plays a crucial role in clinical workflows.  Current methods treat it as a registration or key-point regression task, which has limitations in accurate matching and can only handle predefined landmarks. Recently, some methods have been introduced to address these limitations. One such method, called SAM, proposes using a dense self-supervised approach to learn a distinct embedding for each point on the CT image and achieving promising results. Nonetheless, SAM may still face difficulties when dealing with structures that have similar appearances but different semantic meanings or similar semantic meanings but different appearances. To overcome these limitations, we propose SAM++, a framework that simultaneously learns appearance and semantic embeddings with a novel fixed-points matching mechanism. We tested the SAM++ framework on two challenging tasks, demonstrating a significant improvement over the performance of SAM and outperforming other existing methods.
\end{abstract}
\section{Introduction}
Medical images like CT and MRI provide detailed information about the internal structure of the body. Identifying key anatomical structures from these images plays a crucial role in the clinical workflow, as it helps with diagnosis, treatment planning, and other related procedures \cite{cai2021deep,quan2022images,zhong2019attention}. However, manual annotation in clinical practice is often tedious and repetitive. Therefore, there is a growing interest in developing automatic methods for identifying and matching these structures.
Currently, there are two main types of methods based on the intra- or inter-patient conditions. In intra-patient cases, doctors need to compare target structure changes in time longitudinal CT scans, which can be achieved by registering the newer scans to the original one.  Both traditional registration methods like DEEDS \cite{heinrich2013mrf} and learning-based methods like VoxelMorph \cite{balakrishnan2018unsupervised} produce great results. However, these methods can cause misalignment and distortion in local regions since the registration process optimizes an overall global objective. This could negatively impact the comparison of desired structures. For inter-patients task, most methods address this task by treating it as a supervised landmark detection problem, which uses human-annotated key points as ground truth for training a deep neural network to predict landmark locations on unseen data \cite{o2018attaining,bier2018x,zhu2022datr}. 
Although these methods have demonstrated good performance, they are limited to only handling predefined landmarks. In clinical practice, it's essential for doctors to be able to compare any desired structure across different scans, given the intrinsic similar structure of the human bodies. 
\begin{figure}[htb]
    \centering
    \includegraphics[scale=0.65]{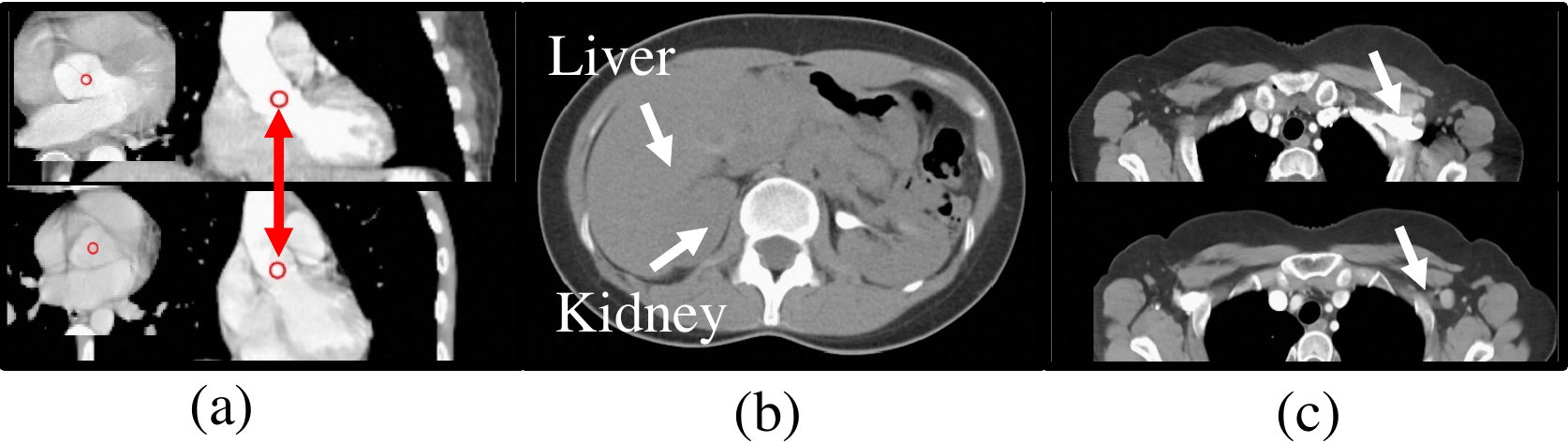}
    \captionsetup{skip=4pt}
    \caption{The diagram of anatomical matching and hard cases. (a)An example shows the matching of aortic valves from two different individuals. (b) In non-contrast CT slices, the adjacent liver and kidney may exhibit very similar texture and intensity, making it difficult to distinguish them using only appearance features. (c) The appearance of the same anatomical structure may vary between contrast-enhanced CT and non-contrast CT scans.}
    \label{fig:fig1}
\end{figure}
Recently, some methods have been introduced to address this requirement \cite{yan2022sam,yao2021one,yao2022relative}. One such method, called SAM \cite{yan2022sam}, proposes using a dense self-supervised approach to learn a distinct embedding for each point on the CT image and perform nearest-neighbor (NN) matching to find the desired structure. This approach has shown promising results in various challenging tasks, including monitoring lesions in longitudinal CT studies \cite{cai2021deep} and matching anatomical structures.  Although this approach has been successful, it may still face limitations when dealing with structures that have \textbf{similar appearances but different semantic meanings} or \textbf{similar semantic meanings but different appearances}. For instance, as shown in Fig. \ref{fig:fig1}, in non-contrast CT slices, the texture of the liver and kidney may appear similar, despite having different semantic meanings. Meanwhile, the use of contrast agents or scans from different times may alter the appearance of the same anatomical structure.

The limitation is due to the self-supervised learning framework used by SAM, which primarily focuses on learning appearance similarities, lacks the ability to recognize higher-level semantic information. Additionally, the nearest-neighbor matching mechanism may also present challenges when the desired structure in the query image is missing or significantly altered.

To address these limitations, we propose SAM++, a framework that simultaneously learns appearance and semantic embeddings combined with a novel fixed-points matching mechanism. Rather than incorporating a semantic segmentation model, SAM++ employs a semantic head on top of the SAM model to generate constant-length semantic embeddings for each point. This has several benefits, as it allows us to concatenate the appearance and semantic embeddings and use them as a unified representation, while also ensuring that the output dimensions are constant. To achieve this, we designed a novel prototypical SupCon loss inspired by the supervised contrastive (SupCon) learning method, enabling contrastive learning in voxel level.  For the matching part, we draw inspiration from the fixed-points method in numerical analysis and propose an iterative method to find high-confidence matchings.  Our method involves starting with matching points from a cube surrounding the template point to the query image, followed by the inverse process from the query image to the template. This process is repeated several times until all points are stabilized in the forward-backward matching. The final query point location is decided by using all these stable points' information.
\setlength{\intextsep}{4pt}
\begin{figure}[tp]
    \centering
    \includegraphics[scale=0.4]{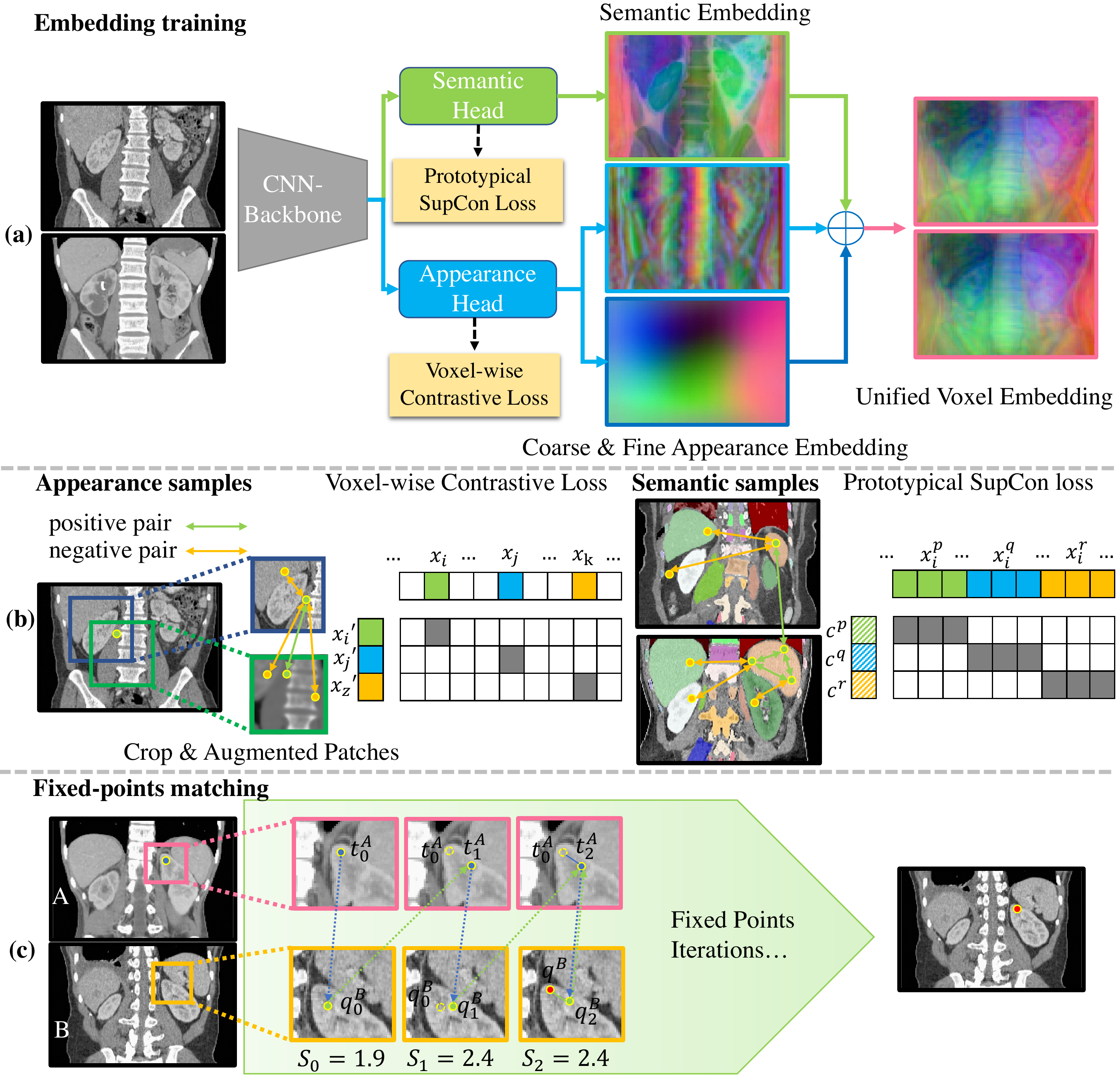}
    \caption{Diagram of the proposed model. }
    \label{fig:framework}
\end{figure}

Our proposed SAM++ framework has been tested on two challenging tasks, namely lesion matching on the DeepLesion DLS dataset \cite{cai2021deep} and Chest CT landmark matching \cite{yan2022sam}. The results demonstrate a significant improvement over the performance of SAM and outperform other existing methods, setting a new state-of-the-art result.

\section{Method}

First, we will present a brief introduction to the SAM model \cite{yan2022sam}, which forms the basis of our method. SAM stands for Self-supervised Anatomical eMbedding and is intended to capture anatomical information for individual voxels in such a way that ensures that similar body parts across images have comparable embeddings. This allows us to identify the same structures in different scans using template-query matching. To achieve this, SAM employs a coarse-to-fine contrastive learning process. At the coarse level, SAM learns the overall global body information, while at the fine level, it learns the precise local information. Given a 3D CT scan, the SAM method first extracts two distinct but partially-overlapping patches from it. These patches are then randomly augmented. By mapping voxels in the overlapped region, we establish correspondences between the locations of one patch and the other. Voxels that appear in the same location on both patches are considered positive pairs, while other voxels on both patches are treated as negative samples. SAM employs hard negative mining to select hard and diverse negative samples from all the negative ones. Finally, the InfoNCE loss is applied to reduce the distance between positive pairs and push positive and negative samples apart. This results in all voxels on the CT scan having a distinct embedding representation and the same location on different augmented views of the same original CT scan having similar embeddings. Due to the high structural similarity of the human body across individuals, SAM can also output similar embeddings for the same anatomical structure on different people's CT scans.

Our SAM++ framework extends SAM and consists of two branches, as shown in Figure \ref{fig:framework}: a semantic branch and an appearance branch. Both branches share the same convolutional neural network (CNN) backbone.  The appearance branch employs the same training method as SAM, where two overlapped and randomly augmented patches  are fed into the CNN-backbone followed by the appearance head to generate appearance embeddings for each voxel. Using $\boldsymbol{x}_i$ and $\boldsymbol{x}_i'$ to represent the embeddings of the positive pair, the appearance branch aims to minimize the voxel-wise contrastive loss, as follows:
\begin{equation}
\mathcal{L}_{app}=-\sum_{i=1}^{n_{\text{pos}}} \log \frac{\exp \left(\boldsymbol{x}_i \cdot \boldsymbol{x}_i' / \tau_{app}\right)}{\exp \left(\boldsymbol{x}_i \cdot \boldsymbol{x}_i' / \tau_{app}\right)+\sum_{j=1}^{n_{\text{neg}}} \exp \left(\boldsymbol{x}_i \cdot \boldsymbol{x}_j / \tau_{app}\right)}.
\end{equation}
Here, $n_{\text{pos}}$ denotes the number of positive pairs, $n_{\text{neg}}$ denotes the number of negative pairs, and $\tau_{app}$ is the temperature parameter.

The self-supervised method utilized by the appearance branch learns to differentiate between two distinct body structures based on their looks. However, this can lead to errors when adjacent tissues or organs share similar intensity and texture, as illustrated in Fig. \ref{fig:fig1}(b). To overcome these challenging cases where appearances are hard to distinguish, we need higher-level semantic differences. Since many CT datasets with organ masks are available, we can directly train another segmentation model and perform segmentation before applying SAM. While this approach is workable, it is inconvenient to use as it doubles the model training and inference time. Additionally, the segmentation model typically regards all regions without a label as background, which is not appropriate since unlabeled body regions also contain various tissues and structures.

Inspired by supervised contrastive learning \cite{khosla2020supervised}, we propose to use a semantic branch to produce a constant-length semantic embedding as a supplement for the appearance embedding. However, using supervised contrastive (SupCon) loss in the voxel level is very expensive, as the complexity of it is $O(n^2)$, where $n$ is the number of voxels. To overcome this difficulty, we design a prototypical SupCon loss by replacing the voxel-voxel positive pairs in SupCon to prototype-voxel pairs. During training, given a batched semantic head output $\{x_i^p\},i\in[1,n_p],p\in[1,K]$ ,where $i$ represents $i_{th}$ voxel embedding with semantic label $p$, we formulate the prototypical SupCon loss as:


\begin{equation}
\mathcal{L}_{sem}=\sum_{p=1}^{K}-\dfrac{1}{n_p}\sum_{i=1}^{n_p}\log \dfrac{\exp{( \boldsymbol{c}_p \cdot  \boldsymbol{x}_i^p/ \tau_{sem})}}{\sum_{o=1}^K\sum_{a=1}^{n_o}\exp{(\boldsymbol{c}_k \cdot \boldsymbol{x}_a^o/ \tau_{sem})}},
\end{equation}
where $\boldsymbol{c}_p=\dfrac{1}{n_p} \sum_{i=1}^{n_p}\boldsymbol{x}_i^p$ is the prototype of class $p$. Compared with the original SupCon loss, prototypical SupCon loss reduces the complexity from $O(n^2)$ to $O(nK)$, enabling its use in dense output tasks.
The output of both appearance and semantic head are normalized vectors, which gives us the benefit of directly concatenate them and use as a single unified embedding.

  We have made progress in improving our embedding representation, to achieve the final goal we also need robust and accurate matching. While the SAM method involves computing the inner product of template and query embeddings and using nearest neighbor (NN) matching, this approach can result in errors when the desired structure in the query image is missing or significantly altered, as we mentioned earlier. 
  Therefore, we propose an alternative method to enhance matching performance. Suppose we have two CT scans, A and B as shown in Fig. \ref{fig:framework}(c), with voxel embeddings of $\boldsymbol{X}^A = \{\boldsymbol{x}_i^A\}$ and $\boldsymbol{X}^B = \{\boldsymbol{x}_i^B \}$, respectively. Given a template embedding on A represented as $\boldsymbol{x}_t^A$, we can find the corresponding query embedding on B by using NN matching: $\boldsymbol{x}_q^B = \mathrm{argmax}_{i\in B}((\boldsymbol{x}_t^A)^T \cdot \boldsymbol{x}_i^B)$. For the sake of simplicity, let's ignore the embedding $\boldsymbol{x}$ and represent the template point and its NN matching point on B as $t_0^A$ and $q_0^B$. We have established that $t_0^A$ corresponds to $q_0^B$, i.e. $t_0^A \rightarrow q_0^B $. Let's now consider the reverse process. Starting from $q_0^B$, can the NN matching method give us $t_0^A$? If it does, we can conclude that we have a consistent forward-backward matching and that is a good matching. However, if the reverse process maps to another point, such as $q_0^B \rightarrow t_1^A$ and $t_1^A \neq t_0^A$, then the first NN match is not reliable. This is because the similarity score of $S_1=q_0^B \rightarrow t_1^A$ is larger than $S_0=q_0^B \rightarrow t_0^A$. Formally, the forward-backward process can be formulated as a function $t_{i+1}^A=f(t_i^A,\boldsymbol{X}^A,\boldsymbol{X}^B)\triangleq t_i^A \rightarrow q_i^B,q_i^B \rightarrow t_{i+1}^A$.

In mathematics, a fixed point of a function is an element that is mapped to itself by the function. Therefore, a forward-backward consistent matching is a fixed point of $f$ since $t_0^A = t_1^A = f(t_0^A)$. For matchings where $t_1^A \neq t_0^A$, although it is not a fixed point, we can always find a fixed point using it as the starting point by fixed-point iteration. For any $t_0^A$, we compute a sequence of its $f$ mappings: ${t_0^A, t_1^A = f(t_0^A), t_2^A = f(t_1^A), t_{i+1}^A = f(t_i^A), ...}$. Eventually, after $n_{fix}$ iterations, the sequence will converge to $t_{i+1}^A = t_i^A, i \geq n_{fix}$. The offset between the starting point and the fixed point is $s_t = t_0^A - t_{n_{fix}}^A$. If $s_t$ is small, we get a nearby fixed point of our template point, and we can use an approximate linear transform $A$ to obtain the query point mapping given by this fixed point: $q^B = q_{n_{fix}}^B - A \cdot s_t$. To compute $A$, we need at least three nearby fixed points. Therefore, we propose searching for fixed points using a cubic region around the template point: first, we select an $L^3$ cubic region centered at the template point $t_0^A$ and perform batched fixed-point iteration.  Then, from the results, we select points with offsets less than a threshold $\tau_{dis}$ to compute $A$ using the least-square estimation. Finally, we get the query point location by averaging the predictions of all the fixed points results. Compared with NN matching, our fixed-points based matching method can adaptively find high-reliable points  and aggregate structural information to give the final matching.  Which is expected to give better matching results.
\section{Experiments}
\subsection{Datasets, Metrics and Implementation Details}

We trained our SAM++ model on two public datasets: NIH-Lymph Node (NIH-LN) \cite{yan2022sam} and the Total Segmentator dataset \cite{wasserthal2022totalsegmentator}. The NIH-LN dataset includes 176 chest-abdomen-pelvis CT scans, while the Total Segmentator dataset contains 1204 CT images with labels of 104 anatomical structures. We evaluated our method on two tasks: lesion tracking and Chest anatomical structure matching. The lesion tracking task aims to match the same lesion on patients' time-longitudinal CT scans. We used the publicly available DLS dataset, which was also used in the DLT \cite{cai2021deep} and TLT \cite{tang2022transformer} methods. The dataset contains 3008, 403, and 480 lesion pairs for training, validation, and testing, respectively. For Chest anatomical structure matching, we used the ChestCT dataset, which was also used in the SAM \cite{yan2022sam} method. The dataset includes 94 patients, each with a contrast-enhanced (CE) and a non-contrast (NC) scan that are pre-aligned.

To evaluate the accuracy of lesion matching, the performance is assessed using the Center Point Matching (CPM) method \cite{tang2022transformer,cai2020deep}. A match is deemed correct if the Euclidean distance between the predicted and ground truth centers is smaller than a threshold. We use 10mm and the lesion radius same as previous methods. The model's evaluation also includes the Mean Euclidean Distance (MED) in mm +/- standard deviation between the predicted and ground truth centers, as well as its projections in each direction (referred to as $\text{MED}_X$, $\text{MED}_Y$, and $\text{MED}_Z$). For the ChestCT matching, we use the same setting as SAM  by calculating the mean distance of 19 predefined landmarks by template-query matching.
\begin{table}
\caption{Lesion tracking comparison on DeepLesion Tracking testing dataset.} \label{tab1}
\resizebox{\textwidth}{!}{
\begin{tabular}{l|c|c|c|c|c|c}
\hline
Method & \tabincell{c}{CPM$@$ \\ 10$mm$} & \tabincell{c}{CPM$@$ \\ Radius} \ & \tabincell{c}{MED$_{X}$ \\ ($mm$)} \ & \tabincell{c}{MED$_{Y}$ \\ ($mm$)} \ & \tabincell{c}{MED$_{Z}$ \\ ($mm$)} \ & \tabincell{c}{MED \\ ($mm$)} \ \\
\hline
Affine~\cite{marstal2016simpleelastix} & $48.33$ & $65.21$ & $4.1\pm5.0$ & $5.4\pm5.6$ & $7.1\pm8.3$ & $11.2\pm9.9$ \\
VoxelMorph~\cite{balakrishnan2018unsupervised} & $49.90$ & $65.59$ & $4.6\pm6.7$ & $5.2\pm7.9$ & $6.6\pm6.2$ & $10.9\pm10.9$ \\
SiamRPN++~\cite{li2019evolution} & $68.85$ & $80.31$ & $3.8\pm4.8$ & $3.8\pm4.8$ & $4.8\pm7.5$ & $8.3\pm9.2$ \\
LENS-LesaNet~\cite{yan2020learning,yan2019holistic} & $70.00$ & $84.58$ & $2.7\pm4.8$ & $2.6\pm4.7$ & $5.7\pm8.6$ & $7.8\pm10.3$ \\
DEEDS~\cite{heinrich2013mrf} & $71.88$ & $85.52$ & $2.8\pm3.7$ & $3.1\pm4.1$ & $5.0\pm6.8$ & $7.4\pm8.1$ \\
DLT-Mix~\cite{cai2021deep} & $78.65$ & $88.75$ & $3.1\pm4.4$ & $3.1\pm4.5$ & $4.2\pm7.6$ & $7.1\pm9.2$ \\
DLT~\cite{cai2021deep} & $78.85$ & $86.88$ & $3.5\pm5.6$ & $2.9\pm4.9$ & $4.0\pm6.1$ & $7.0\pm8.9$ \\
TransT~\cite{chen2021transformer} & $79.59$ & $88.99$ & $3.4\pm5.9$ & $5.4\pm6.1$ & $1.8\pm2.2$ & $7.6\pm7.9$ \\
SAM \cite{yan2022sam} & $86.04$ & $95.00$ & $2.6\pm3.8$ & $2.3\pm2.9$ & $4.0\pm5.6$ & $6.1\pm6.7$ \\
TLT \cite{tang2022transformer} & $87.37$ & $95.32$ & $3.0\pm6.2$ & $3.7\pm5.2$ & $\mathbf{1.7\pm2.1}$ & $6.1\pm6.7$ \\
\hline
SAM++ & $\mathbf{88.33}$ & $\mathbf{96.35}$ & $\mathbf{2.4\pm3.2}$ & $\mathbf{2.1\pm2.6}$ & $3.8\pm5.3$ & $\mathbf{5.4\pm6.0}$ \\
\hline
\end{tabular}
}
\end{table}

The proposed method is implemented using PyTorch (v1.9) and MMDetction (v1.20). Same to SAM, we use 3D ResNet18 as our CNN-backbone and 3D feature pyramid network (FPN) as our semantic and appearance head. The embedding length of each head is set as 128. For the appearance head, we output a coarse level and a fine level embedding same as SAM. For the semantic head, we only output the fine-level embedding. To save GPU memory, the size of the model's output is half of the input volume size, we then use trilinear interpolation to recover it to the original input size.
The network is optimized by SGD with momentum=0.9 and the learning rate is set to 0.02. The batch size is 5 and temperature  $\tau_{app}$ and $\tau_{sem}$ are all set to 0.5. All CT volumes have been resampled to the isotropic resolution of 2mm. For fixed-points matching, we set $L=5$.We use random rotation, random resample (scale in $[0.8,1.2]$), random noise, and random blur for data augmentation.

\subsection{Experimental Results and Discussion}

Table \ref{tab1} displays the results of the DLS test set. The comparison methods can be categorized based on whether they use lesion annotations. Methods such as DLT\cite{cai2021deep}, TransT\cite{chen2021transformer}, and TLT \cite{tang2022transformer} utilize lesion annotations during training and therefore have task-specific supervision. Conversely, other methods such as VoxelMorph\cite{balakrishnan2018unsupervised}, DEEDS \cite{heinrich2013mrf}, and SAM \cite{yan2022sam} do not use any lesion information. As we can see, our SAM++ method outperformed all other methods. To provide a more comprehensive analysis, since we do not use DLS training set in model training, we conducted ablation on the full DSL dataset, which includes 3883 pairs of lesions. The results are presented in Table \ref{tab2}, and we note that the SAM method itself yields a robust performance. By incorporating the semantic branch, we enhance the CPM@Radius from 93.44 to 93.99. Additionally, by employing structural inference techniques on SAM, we achieve a performance boost to 94.44. These results demonstrate that both our contributions are effective. Finally, by combining these two techniques, our SAM++ model can further increase the CPM@Radius to 95.45. For Chest CT dataset, we validate our method on all 94 cases and tested both intra- and inter- phases matching, as shown in Table \ref{tab3} our method achieves best results on all settings.

\begin{table}
\caption{Abalation results on full DLS dataset.} \label{tab2}
\resizebox{\textwidth}{!}{
\begin{tabular}{l|c|c|c|c|c|c}
\hline
Method & \tabincell{c}{CPM$@$ \\ 10$mm$} & \tabincell{c}{CPM$@$ \\ Radius} \ & \tabincell{c}{MED$_{X}$ \\ ($mm$)} \ & \tabincell{c}{MED$_{Y}$ \\ ($mm$)} \ & \tabincell{c}{MED$_{Z}$ \\ ($mm$)} \ & \tabincell{c}{MED \\ ($mm$)} \ \\
\hline
SAM & $88.94$ & $93.44$ & $2.4\pm 3.0$ & $2.5\pm3.1$ & $3.6\pm3.8$ & $5.8\pm5.0$ \\
SAM+Semantic & $89.37$ & $93.99$ & $2.3\pm2.6$ & $2.5\pm2.8$ & $3.5\pm3.7$ & $5.6\pm4.5$ \\
SAM+Structual & $89.66$ & $94.44$ & $2.2\pm2.4$ & $2.3\pm2.8$ & $3.4\pm3.7$ & $5.3\pm4.4$ \\
SAM++ & $\mathbf{91.05}$ & $\mathbf{95.45}$ & $\mathbf{2.2\pm2.2}$ & $\mathbf{2.1\pm2.2}$ & $\mathbf{3.2\pm3.4}$ & $\mathbf{5.1\pm3.8}$ \\
\hline
\end{tabular}
}
\end{table}

\begin{table}
\caption{Comparison of methods on the ChestCT dataset.} \label{tab3}
\resizebox{\textwidth}{!}{
\begin{tabular}{l|c|c|c|c}
\hline
Method & \tabincell{c}{CE-CE} & \tabincell{c}{NC-NC} \ & \tabincell{c}{CE-NC} \ & \tabincell{c}{NC-CE} \\
\hline
 \multicolumn{5}{c}{On 19 test cases} \\
\hline
Affine~\cite{marstal2016simpleelastix} ~~~~~ & 8.4$\pm$5.2~32.9 & 8.5$\pm$5.3~33.1&-&-\\\hline
DEEDS~\cite{heinrich2013mrf} &4.6$\pm$3.3~18.8& 4.7$\pm$3.4~24.4&-&-\\\hline
VoxelMorph~\cite{balakrishnan2018unsupervised}  &7.3$\pm$3.6~20.1&7.4$\pm$3.7~20.2&-&-\\\hline
SAM &4.3$\pm$3.0~16.4&4.5$\pm3.0$~18.5&-&-\\
\hline
 \multicolumn{5}{c}{On all 94 cases} \\
\hline
SAM & 4.8$\pm$3.4~23.6&4.6$\pm$3.5~25.9&4.7$\pm$3.5~44.0&5.2$\pm$3.9 29.6\\\hline
SAM++ &\textbf{4.0$\pm$2.5~16.1}&\textbf{3.9$\pm$2.4~17.0}&\textbf{4.0$\pm$2.5~17.2}&\textbf{4.0$\pm$2.5~16.9}\\\hline
\end{tabular}
}
\end{table}
\section{Conclusion}

%
%
%
\bibliographystyle{splncs04}
\bibliography{ijcai22}

\begin{thebibliography}{10}
\providecommand{\url}[1]{\texttt{#1}}
\providecommand{\urlprefix}{URL }
\providecommand{\doi}[1]{https://doi.org/#1}

\bibitem{balakrishnan2018unsupervised}
Balakrishnan, G., Zhao, A., Sabuncu, M.R., Guttag, J., Dalca, A.V.: An
  unsupervised learning model for deformable medical image registration. In:
  Proceedings of the IEEE conference on computer vision and pattern
  recognition. pp. 9252--9260 (2018)

\bibitem{bier2018x}
Bier, B., Unberath, M., Zaech, J.N., Fotouhi, J., Armand, M., Osgood, G.,
  Navab, N., Maier, A.: X-ray-transform invariant anatomical landmark detection
  for pelvic trauma surgery. In: Medical Image Computing and Computer Assisted
  Intervention--MICCAI 2018: 21st International Conference, Granada, Spain,
  September 16-20, 2018, Proceedings, Part IV. pp. 55--63. Springer (2018)

\bibitem{cai2021deep}
Cai, J., Tang, Y., Yan, K., Harrison, A.P., Xiao, J., Lin, G., Lu, L.: Deep
  lesion tracker: monitoring lesions in 4d longitudinal imaging studies. In:
  Proceedings of the IEEE/CVF Conference on Computer Vision and Pattern
  Recognition. pp. 15159--15169 (2021)

\bibitem{cai2020deep}
Cai, J., Yan, K., Cheng, C.T., Xiao, J., Liao, C.H., Lu, L., Harrison, A.P.:
  Deep volumetric universal lesion detection using light-weight pseudo 3d
  convolution and surface point regression. In: International Conference on
  Medical Image Computing and Computer-Assisted Intervention. pp. 3--13.
  Springer (2020)

\bibitem{chen2021transformer}
Chen, X., Yan, B., Zhu, J., Wang, D., Yang, X., Lu, H.: Transformer tracking.
  In: Proceedings of the IEEE/CVF Conference on Computer Vision and Pattern
  Recognition. pp. 8126--8135 (2021)

\bibitem{heinrich2013mrf}
Heinrich, M.P., Jenkinson, M., Brady, M., Schnabel, J.A.: Mrf-based deformable
  registration and ventilation estimation of lung ct. IEEE transactions on
  medical imaging  \textbf{32}(7),  1239--1248 (2013)

\bibitem{khosla2020supervised}
Khosla, P., Teterwak, P., Wang, C., Sarna, A., Tian, Y., Isola, P., Maschinot,
  A., Liu, C., Krishnan, D.: Supervised contrastive learning. Advances in
  neural information processing systems  \textbf{33},  18661--18673 (2020)

\bibitem{li2019evolution}
Li, B., Wu, W., Wang, Q., Zhang, F., Xing, J., Yan, J.S., et~al.: Evolution of
  siamese visual tracking with very deep networks. In: Proceedings of the IEEE
  Conference on Computer Vision and Pattern Recognition, Long Beach, CA, USA.
  pp. 16--20 (2019)

\bibitem{marstal2016simpleelastix}
Marstal, K., Berendsen, F., Staring, M., Klein, S.: Simpleelastix: A
  user-friendly, multi-lingual library for medical image registration. In:
  Proceedings of the IEEE conference on computer vision and pattern recognition
  workshops. pp. 134--142 (2016)

\bibitem{o2018attaining}
O'Neil, A.Q., Kascenas, A., Henry, J., Wyeth, D., Shepherd, M., Beveridge, E.,
  Clunie, L., Sansom, C., Seduikyte Keith~Muir, E., Poole, I.: Attaining
  human-level performance with atlas location autocontext for anatomical
  landmark detection in 3d ct data. In: Proceedings of the European Conference
  on Computer Vision (ECCV) Workshops. pp.~0--0 (2018)

\bibitem{quan2022images}
Quan, Q., Yao, Q., Li, J., Zhou, S.K.: Which images to label for few-shot
  medical landmark detection? In: Proceedings of the IEEE/CVF Conference on
  Computer Vision and Pattern Recognition. pp. 20606--20616 (2022)

\bibitem{tang2022transformer}
Tang, W., Kang, H., Zhang, H., Yu, P., Arnold, C.W., Zhang, R.: Transformer
  lesion tracker. In: Medical Image Computing and Computer Assisted
  Intervention--MICCAI 2022: 25th International Conference, Singapore,
  September 18--22, 2022, Proceedings, Part VI. pp. 196--206. Springer (2022)

\bibitem{wasserthal2022totalsegmentator}
Wasserthal, J., Meyer, M., Breit, H.C., Cyriac, J., Yang, S., Segeroth, M.:
  Totalsegmentator: robust segmentation of 104 anatomical structures in ct
  images. arXiv preprint arXiv:2208.05868  (2022)

\bibitem{yan2022sam}
Yan, K., Cai, J., Jin, D., Miao, S., Guo, D., Harrison, A.P., Tang, Y., Xiao,
  J., Lu, J., Lu, L.: Sam: Self-supervised learning of pixel-wise anatomical
  embeddings in radiological images. IEEE Transactions on Medical Imaging
  \textbf{41}(10),  2658--2669 (2022)

\bibitem{yan2020learning}
Yan, K., Cai, J., Zheng, Y., Harrison, A.P., Jin, D., Tang, Y., Tang, Y.,
  Huang, L., Xiao, J., Lu, L.: Learning from multiple datasets with
  heterogeneous and partial labels for universal lesion detection in ct. IEEE
  Transactions on Medical Imaging  \textbf{40}(10),  2759--2770 (2020)

\bibitem{yan2019holistic}
Yan, K., Peng, Y., Sandfort, V., Bagheri, M., Lu, Z., Summers, R.M.: Holistic
  and comprehensive annotation of clinically significant findings on diverse ct
  images: learning from radiology reports and label ontology. In: Proceedings
  of the IEEE/CVF Conference on Computer Vision and Pattern Recognition. pp.
  8523--8532 (2019)

\bibitem{yao2021one}
Yao, Q., Quan, Q., Xiao, L., Kevin~Zhou, S.: One-shot medical landmark
  detection. In: Medical Image Computing and Computer Assisted
  Intervention--MICCAI 2021: 24th International Conference, Strasbourg, France,
  September 27--October 1, 2021, Proceedings, Part II 24. pp. 177--188.
  Springer (2021)

\bibitem{yao2022relative}
Yao, Q., Wang, J., Sun, Y., Quan, Q., Zhu, H., Zhou, S.K.: Relative distance
  matters for one-shot landmark detection. arXiv preprint arXiv:2203.01687
  (2022)

\bibitem{zhong2019attention}
Zhong, Z., Li, J., Zhang, Z., Jiao, Z., Gao, X.: An attention-guided deep
  regression model for landmark detection in cephalograms. In: Medical Image
  Computing and Computer Assisted Intervention--MICCAI 2019: 22nd International
  Conference, Shenzhen, China, October 13--17, 2019, Proceedings, Part VI 22.
  pp. 540--548. Springer (2019)

\bibitem{zhu2022datr}
Zhu, H., Yao, Q., Zhou, S.K.: Datr: Domain-adaptive transformer for
  multi-domain landmark detection. arXiv preprint arXiv:2203.06433  (2022)

\end{thebibliography}
%




\end{document}